\documentclass[journal]{IEEEtran}
\usepackage{graphicx}
\usepackage{amssymb}
\usepackage{amsthm}
\usepackage{amsmath}
\usepackage{xcolor}
\usepackage{caption}
\usepackage{subcaption}
\usepackage[noend,ruled,vlined]{algorithm2e}
\usepackage{footnote}

\def\eg{\emph{e.g., }}
\def\ie{\emph{i.e., }}
\def\etal{\emph{et al.}}

\theoremstyle{remark}
\newtheorem{remark}{Remark}

\begin{document}
\title{DAE-PINN:~A Physics-Informed Neural Network Model for Simulating Differential Algebraic Equations with Application to Power Networks}
\author{Christian~Moya,~\IEEEmembership{Member,~IEEE,}
        and~Guang~Lin,~\IEEEmembership{Senior~Member,~IEEE}
\thanks{C. Moya and G. Lin are with the Department of Mathematics, Purdue University, West Lafayette,
IN, 47907 USA e-mail: \{cmoyacal, guanglin\}@purdue.edu.}}
\markboth{Journal of }%
{Shell \MakeLowercase{\textit{et al.}}: Bare Demo of IEEEtran.cls for IEEE Journals}
% abstract
\maketitle
\begin{abstract}
Deep learning-based surrogate modeling is becoming a promising approach for learning and simulating dynamical systems. Deep-learning methods, however, find very challenging learning stiff dynamics. In this paper, we develop DAE-PINN, the first effective deep-learning framework for learning and simulating the solution trajectories of \textit{nonlinear differential-algebraic equations}~(DAE), which present a form of \textit{infinite stiffness} and describe, for example, the dynamics of power networks. Our DAE-PINN bases its effectiveness on the synergy between \textit{implicit Runge-Kutta} time-stepping schemes (designed specifically for solving DAEs) and \textit{physics-informed neural networks}~(PINN) (deep neural networks that we train to satisfy the dynamics of the underlying problem). Furthermore, our framework (i) enforces the neural network to satisfy the DAEs as (approximate) hard constraints using a penalty-based method and (ii) enables simulating DAEs for long-time horizons. We showcase the effectiveness and accuracy of DAE-PINN by learning and simulating the solution trajectories of a three-bus power network.
\end{abstract}
% keywords
\IEEEpeerreviewmaketitle
% introduction
\section{Introduction}
\IEEEPARstart{I}{n} recent years, we have seen the power network incorporate more and more transformative technologies such as integrating distributed energy resources, enabling a liberalized market, or adopting more complex communication and control algorithms. Such transformation seeks to enhance the reliability and efficiency of the power network operation. This transformation, however, pushes the power network to operate under a more diversified set of operating conditions and contingencies that could potentially compromise its security.  

To assess the power network's dynamic security~\cite{kundur2007power}, operators implement an offline procedure that seeks to predict whether the power network will remain safely operating after facing a single contingency (e.g., the disconnection of a generator) from a set of credible contingencies. Such a procedure is known as the $N-1$ criteria~\cite{alvarado2002transmission} and requires simulating the power network's dynamic response.

Simulating the power network's dynamic response requires integrating a set of nonlinear differential-algebraic equations~(DAE)~\cite{kundur2007power}. Solving this set of DAEs is, however, a challenging task. Indeed, the classical explicit integration schemes fail catastrophically on such a task~\cite{iserles2009first}. As a result, most commercial solvers for DAEs use numerically stable schemes to integrate the dynamic equations and iterative schemes to solve the algebraic equations~\cite{stott1979power}. However, the computational cost and memory required to integrate DAEs are very high and constitute the main obstacle to deploying dynamic security assessment online~\cite{schainker2006real}. However, with the transformation the power network now faces, soon, it will become imperative for electric utilities to assess security online, which calls for the faster integration/simulation of DAEs.

Motivated by the above power network application, in this paper, we seek to derive a deep learning~(DL) framework that accelerates simulating nonlinear DAEs. Enabled by the exponential growth of computational power and data availability, DL has achieved outstanding performance in computer vision and natural language processing applications~\cite{lecun2015deep}, and promises to also revolutionize the scientific and engineering fields. However, the current application of DL to learn scientific and engineering dynamical systems is, at most, limited since the cost of collecting data is prohibitive. Most conventional DL methods (e.g., convolutional or recurrent neural networks) lack robustness and generalization capabilities in such a small data regime.

In recent years, the field of scientific machine learning~\cite{baker2019workshop} has provided us with a series of new transformative works~\cite{raissi2019physics,lu2021deepxde, lu2021physics, yazdani2020systems} aiming at learning the differential equations describing dynamical systems and, hence providing us with an efficient alternative to traditional costly numerical solvers. Behind most of these transformative works lies the idea of using the physical laws that govern these dynamical systems~\cite{raissi2019physics}. Such prior information acts as a regularizing agent, limiting the space of possible solutions and enabling generalizing well even when the amount of data is small. Admittedly, there is still much work needed to scale physics-informed deep learning methods so that they can become \textit{accurate} surrogate models of large-scale dynamical systems. In particular, these accurate surrogate models must (i) predict solution trajectories for a large set of initial conditions and (ii) maintain physical accuracy for long-time horizons.

Despite the success of scientific machine learning for learning the solution trajectories of ordinary differential equations~\cite{yazdani2020systems}, developing a DL-based framework for learning and simulating the solution trajectories of nonlinear differential-algebraic equations remains an open problem. This is because DAEs present a form of infinite stiffness~\cite{kim2021stiff} that may produce gradient pathologies~\cite{wang2020understanding} and ill-conditioned optimization problems, leading to the failure of the stochastic gradient descent-based training. The first attempts to derive DL frameworks for learning stiff differential equations were presented in~\cite{ji2020stiff} and \cite{kim2021stiff}. In~\cite{ji2020stiff}, the authors show that continuous PINN models fail to learn stiff ODEs and propose using quasi-steady-state assumptions to derive a simpler model more suitable for PINNs. In~\cite{kim2021stiff}, Kim \etal~modified neural ordinary differential equations~\cite{chen2018neural} so that they can learn the solution trajectories of stiff problems for long-time horizons. Both of the above methods have their merits, but, as presented, they are not suitable for learning the solution trajectories of the DAEs studied in this paper.

In this paper, we develop DAE-PINN, the first deep learning-based framework for learning and simulating the solution trajectories of semi-explicit \textit{differential-algebraic equations}~(DAE) of index-1. In particular, our objectives in this paper are:
\begin{enumerate}
    \item \textit{Forward problem}: deriving a framework that learns to map a given distribution of initial conditions to the solution trajectories (within a short-time interval) of a dynamical system described by DAEs.
    \item \textit{Long-time simulation of DAEs}: designing an algorithm that uses the trained framework to simulate DAEs over long-time horizons.
\end{enumerate}
We detail our contributions next.
\begin{enumerate}
    \item We design a deep learning~(DL) framework (DAE-PINN - Sections~\ref{ssec:IRK} and \ref{ssec:discrete-PINNs}) that tackles the \textit{forward problem} by enabling the synergistic combination of a discrete \textit{physics-informed neural network} model with an \textit{implicit Runge-Kutta} scheme designed specifically for solving DAEs. Thus, our framework effectively extends the method proposed in~\cite{raissi2019physics} to DAEs.
    \item A \textit{penalty}-based method is then introduced (Section~\ref{ssec:penalty-method}) to facilitate the training of DAE-PINN. The penalty method aims to enforce DAE-PINN to satisfy the DAEs as (approximate) hard constraints.
    \item For the \textit{long-time simulation of DAEs}, we propose an algorithm (Section~\ref{ssec:simulating-DAEs}) that iteratively evaluates the trained DAE-PINN. Following a Markov-like procedure, the proposed algorithm uses the DAE-PINN prediction of the previous evaluation step as the initial condition for the next step. 
    \item We illustrate the training protocols for DAE-PINN and evaluate its effectiveness (Section~\ref{sec:numerical-experiments}) using a three-bus power network example described by a set of stiff and nonlinear DAEs.
\end{enumerate}
We organize this work as follows. In Section~\ref{sec:problem-description}, we introduce the \textit{differential algebraic equations}~(DAE) studied in this paper. In Section~\ref{sec:proposed-method}, after describing the implicit Runge-Kutta~(IRK) time-stepping scheme, we describe DAE-PINN, \ie the discrete \textit{physics-informed neural network} that allows us to use the IRK scheme (with an arbitrary number of stages) for solving DAEs. We then describe the penalty method that enforces DAE-PINN to satisfy the DAEs as approximate hard constraints. We conclude Section~\ref{sec:proposed-method} by introducing Algorithm~\ref{alg:integrating-long-time-horizons} that enables us to use the trained DAE-PINN for simulating DAEs over long-time horizons. In Section~\ref{sec:numerical-experiments}, we verify the effectiveness of the proposed framework using a three-bus power network example. We provide a discussion of our results and future work in Section~\ref{sec:discussion} and conclude the paper in Section~\ref{sec:conclusion}.
%
%% problem setup
\section{Problem setup}
\label{sec:problem-description}
In this paper, we develop DAE-PINN, a deep learning-based framework that employs physics-informed neural networks~\cite{raissi2019physics} and implicit Runge-Kutta schemes~\cite{iserles2009first} for learning the solution trajectories of nonlinear \textit{Differential-Algebraic equations}~(DAE)~\cite{wanner1996solving} given in the semi-explicit form
\begin{subequations} \label{eq:DAE}
\begin{align}
    \dot{y} &= f(y,z), \qquad y(t_0) = y_0 \label{eq:dynamic} \\
    0 &=g(y,z), \qquad z(t_0) = z_0, \label{eq:algebraic}
\end{align}
\end{subequations}
where $y = y(t) \in \mathbb{R}^n$ are the dynamic states, $z = z(t) \in \mathbb{R}^m$ are the algebraic variables, $f:\mathbb{R}^n \times \mathbb{R}^m \to \mathbb{R}^n$ describes the differential equations,  $g:\mathbb{R}^n \times \mathbb{R}^m \to \mathbb{R}^m$ the algebraic equations, $t \in [t_0, T]$ the simulation time interval, and $T > t_0$ the time horizon.
\\~\\
\noindent \textit{Assumptions:} Let us assume that $f$ and $g$ are sufficiently often differentiable and the initial conditions satisfy $g(y_0,z_0) = 0$. We also assume that the DAEs~\eqref{eq:DAE} are of index 1~\cite{roche1989implicit}, which means that the inverse of the Jacobian $g_z = \partial g/ \partial z$ exists and is bounded in a neighborhood of the exact solution. This implies that, by the implicit function theorem~\cite{rudin1976principles}, the algebraic equations~\eqref{eq:algebraic} have locally a unique solution $z = G(y)$. Hence the DAE~\eqref{eq:DAE} is equivalent to the following system of ordinary differential equations
\begin{align} \label{eq:ode}
    \dot{y} = f(y, G(y)).
\end{align}
with initial conditions $(y(t_0),z(t_0)) = (y_0,G(y_0))$. Notice that the examples studied in Ji \etal~\cite{ji2020stiff} (Stiff-PINNs) correspond to a special case in our problem setup where the algebraic variables~$z$ can be solved for explicitly to obtain~\eqref{eq:ode}. 
\\~\\
\textit{Applications:} DAEs frequently arise in dynamic simulations of power networks~\cite{kundur2007power}, mechanical problems, trajectory control, etc. DAEs also originate from singular perturbation problems~(SPP) of the form
\begin{subequations} \label{eq:spp}
\begin{align}
    \dot{y} &= f(y,z) \\
    \epsilon \dot{z} &=g(y,z),
\end{align}
\end{subequations}
by letting the parameter~$\epsilon > 0$ approach zero. SPPs have been used to study (i) nonlinear oscillations with large parameters, (ii) structure-preserving power networks with frequency-dependent dynamic loads, and (iii) chemical kinetics with slow and fast reactions.

We conclude this section with the following remark. The DAE-PINN that we will develop in Section~\ref{sec:proposed-method} could also be used for solving problems described in descriptor form   
\begin{align}  \label{eq:descriptor}
    M \dot{x} = \varphi(x), \qquad x(t_0) = x_0,
\end{align}
where $x \in \mathbb{R}^{n+m}$ and $M$ is a singular matrix. To that end, we show next that~\eqref{eq:descriptor} is mathematically equivalent to the DAE~\eqref{eq:DAE}. First, we decompose~$M$ (\eg via Gaussian elimination with total pivoting) as
$$
M = S \begin{pmatrix} I & 0 \\ 0 & 0 \end{pmatrix} T
$$
where $S$ and $T$ are invertible matrices and $I$ is the identity matrix with dimension corresponding to the rank of $M$. Then, we insert the above into~\eqref{eq:descriptor} and use $Tu = \left(y^\top,z^\top \right)^\top$ to obtain
$$
    \begin{pmatrix} \dot{y} \\ 0 \end{pmatrix} = S^{-1} \varphi \left( T^{-1} \begin{pmatrix} y \\ z \end{pmatrix}  \right) =: \begin{pmatrix} f(y,z) \\ g(y,z) \end{pmatrix},
$$
\ie the semi-explicit DAE~\eqref{eq:DAE}. Thus, the deep learning framework that we will derive in Section~\ref{sec:proposed-method} for~\eqref{eq:DAE} also applies for problems in descriptor form~\eqref{eq:descriptor}, provided we can decompose the matrix~$M$.

%% proposed method
\section{Proposed method - DAE-PINN}
\label{sec:proposed-method}
This section describes our DAE-PINN framework, \ie a physics-informed neural network framework that allows to solve the DAE~\eqref{eq:DAE} using the \textit{implicit Runge-Kutta}~(IRK) time-stepping scheme with $\nu$ stages. 

\subsection{Implicit Runge-Kutta Scheme}
\label{ssec:IRK}
Let us start by assuming that the integration of~\eqref{eq:DAE} has been carried out up to~$(t_n, y_n, z_n)$ and we seek to advance it to~$(t_{n+1}, y_{n+1}, z_{n+1})$, where $t_{n+1} = t_n + h$ and $h > 0$ is the \textit{time step}~\cite{iserles2009first}. We apply the implicit Runge-Kutta scheme with $\nu$ stages~\cite{iserles2009first,roche1989implicit} to our system of DAEs~\eqref{eq:DAE} and obtain
\begin{subequations} \label{eq:IRK-DAE}
\begin{align}
    \xi_j &= y_n + h \sum_{i=1}^{\nu} a_{j,i} f(\xi_i,\zeta_i),~\quad j=1,\ldots,\nu \\
    0 &= g(\xi_j,\zeta_j),~\qquad~\qquad~\qquad~j=1,\ldots,\nu \\
    y_{n+1} &= y_n + h \sum_{j=1}^{\nu} b_{j} f(\xi_j,\zeta_j) \\
    0 &= g(y_{n+1},z_{n+1}).
\end{align}
\end{subequations}
Here $\xi_j = y(t_n + c_j h)$, $\zeta_j = z(t_n + c_j h)$, and $\{a_{j,i},b_j,c_i\}$ are the known parameters of the IRK scheme. Following~\cite{iserles2009first} and to let the scheme be of nontrivial order, we impose the following convention for the parameters
$$
\sum_{i = 1}^\nu a_{j,i} = c_j.
$$

\subsection{Discrete Physics-Informed Neural Networks}
\label{ssec:discrete-PINNs}
In classical numerical analysis~\cite{iserles2009first}, implicit formulations of Runge-Kutta schemes are usually constrained due to the computational cost of solving~\eqref{eq:IRK-DAE}. And if we increase the number of IRK stages, these constraints become more severe. To overcome these constraints, our DAE-PINN framework employs a \textit{discrete physics-informed neural network}~(PINN) model~\cite{raissi2019physics} to enable the implicit Runge-Kutta scheme with $\nu$ stages~\eqref{eq:IRK-DAE}.

\begin{figure}
     \centering
     \begin{subfigure}[b]{0.5\textwidth}
         \centering
         \includegraphics[width=.7\textwidth, height=4.35cm]{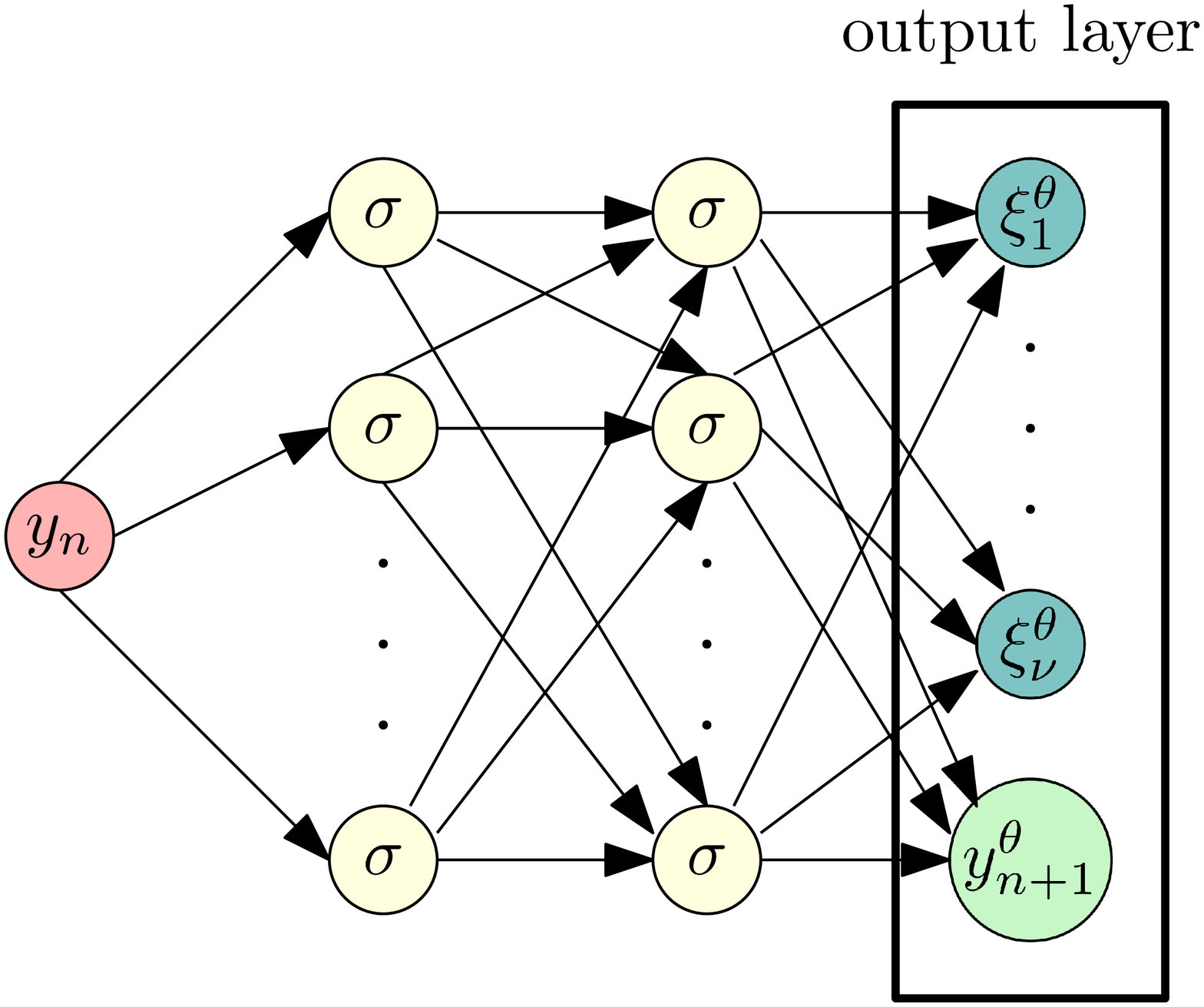}
         \caption{}
         \label{fig:discrete-PINN}
     \end{subfigure}
     \hfill
     \begin{subfigure}[b]{0.5\textwidth}
         \centering
         \includegraphics[width=.60\textwidth, height=4.25cm]{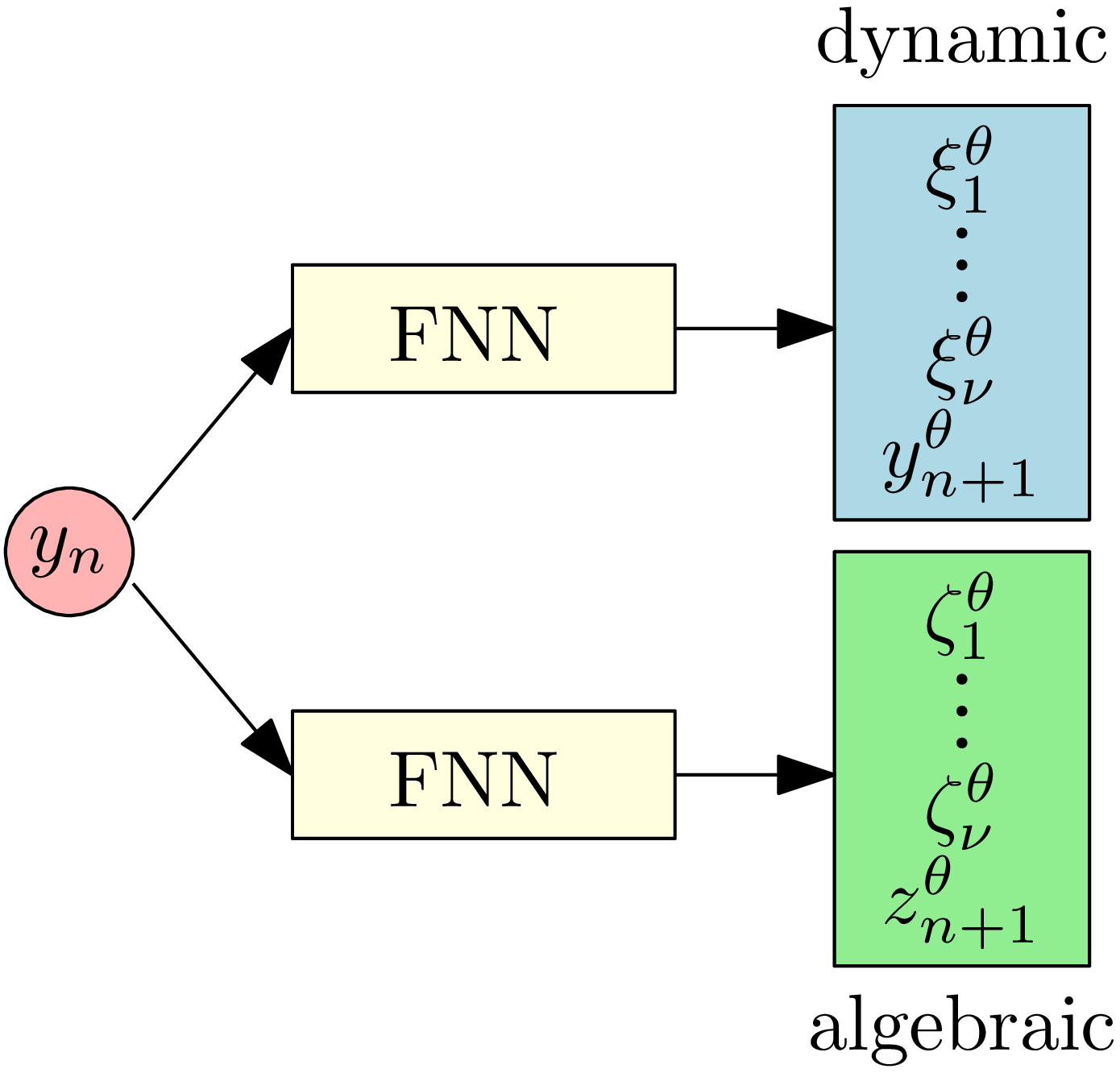}
         \caption{}
         \label{fig:stacked}
     \end{subfigure}
        \caption{(a)~The multi-output fully-connected neural network for the dynamic states~$y$. (b)~Unstacked architecture - DAE-PINN framework for solving the DAEs~\eqref{eq:DAE} using IRK~\eqref{eq:IRK-DAE}.}
        \label{fig:nth-DAE-PINN}
        \vspace{-1em}
\end{figure}

In the discrete PINN model, the first step is to construct multi-output neural networks with parameters~$\theta$~(see Fig.~\ref{fig:discrete-PINN} and \ref{fig:stacked}) as a surrogate for the solution of the IRK scheme~\eqref{eq:IRK-DAE}, which takes the input~$y_n$ and outputs

\begin{subequations} \label{eq:discrete-PINN}
\begin{align}
    &[\xi^\theta_1, \ldots, \xi^\theta_\nu, y^\theta_{n+1}] \\
    &[\zeta^\theta_1,\ldots,\zeta^\theta_\nu,z^\theta_{n+1}].
\end{align}
\end{subequations}
\begin{remark}
\textit{Architecture:} In this paper, we mainly adopt the \textit{unstacked} architecture depicted in Fig.~\ref{fig:stacked}, which assigns one neural network for the dynamic state variables~$y \in \mathbb{R}^n$ and another neural network for the algebraic variables~$z \in \mathbb{R}^m$. We remark, however, that one can also adopt a \textit{stacked} architecture, which assigns a single neural network for each of the dynamic variables~$y_i \in \mathbb{R}$ and each of the algebraic variables~$z_i \in \mathbb{R}$. 
\end{remark}

\begin{remark}
\textit{Complexity:} As described in~\cite{raissi2019physics}, PINN enable us to employ implicit Runge-Kutta schemes with a large number of stages at effectively very little extra cost. More specifically, only the number of neurons of the last layer of the neural networks grows linearly with the total number of stages, \ie with cost~$\sim O(\nu)$. 
\end{remark}

In the second step for the discrete PINN model, we restrict the neural networks to satisfy the differential and algebraic equations described by the IRK scheme~\eqref{eq:IRK-DAE}. In practice, we restrict the neural networks on some set of randomly distributed initial conditions scattered/sampled throughout the domain~\cite{lu2021deepxde}, \ie the set of training data $\mathcal{T}:=\{y_n^1,y_n^2,\hdots,y_n^{|\mathcal{T}|}\}$ of size~$|\mathcal{T}|$\footnote{Observe that our proposed framework does not require supervision, \ie it does not require to know target values of the solution trajectory.}. To measure the discrepancy between the neural networks and the IRK scheme~\eqref{eq:IRK-DAE}, we use the following loss function:
\begin{align} \label{eq:loss-fn}
    \mathcal{L}(\theta;\mathcal{T}) = w_f \mathcal{L}_f(\theta;\mathcal{T}) + w_g \mathcal{L}_g(\theta;\mathcal{T}).
\end{align}
In the above, $w_g$ and $w_f$ are the weights,
$$
\mathcal{L}_f(\theta;\mathcal{T}) = \frac{1}{|\mathcal{T}|(\nu+1)} \sum_{y_n \in \mathcal{T}} \sum_{j=1}^{\nu+1} ||y_n - y^n_j(\theta)||_2^2,
$$
where
\begin{align*}
    y_j^n(\theta) &:= \xi^\theta_k - h \sum_{i=1}^\nu a_{j,i}f(\xi^\theta_i,\zeta^\theta_i),~\quad j=1,\ldots,\nu \\
    y_{\nu+1}^n (\theta) &:= y^\theta_{n+1} - h \sum_{j=1}^s b_{j}f(\xi^\theta_j,\zeta^\theta_j),
\end{align*}
and
$$
\mathcal{L}_g(\theta;\mathcal{T}) = \frac{1}{(\nu+1)} \left(\sum_{j=1}^{\nu} ||g(\xi^\theta_j,\zeta^\theta_j)||_2^2 + ||g(y^\theta_{n+1},z^\theta_{n+1})||_2^2 \right).
$$

In the last step for the discrete PINN model, we train the neural network parameters by minimizing the loss function using gradient-based optimizers, \eg the Adam optimizer~\cite{kingma2014adam}:
\begin{align} \label{eq:optim}
    \theta^* = \text{arg}\min_\theta \mathcal{L}(\theta;\mathcal{T}).
\end{align}
We use the weight coefficients $w_f$ and $w_g$ in~\eqref{eq:loss-fn} to balance the residual loss terms for the dynamic variables~$\mathcal{L}_f$ and the algebraic variables~$\mathcal{L}_g$. In this paper, we use a \textit{penalty}-based method~\cite{lu2021physics} to update the value of the weight coefficients $w_f$ and $w_g$. 
\vspace{-1em}
\subsection{Enforcing DAEs as approximate hard constraints}
\label{ssec:penalty-method}
In the DAE problem~\eqref{eq:DAE}, the solution trajectories must always satisfy the dynamic equations~\eqref{eq:dynamic} and lie in the manifold described by the algebraic equations~\eqref{eq:algebraic}, \ie
$$
\{(y,z)~:~g(y,z) = 0\}.
$$
which, for power networks, represents satisfying the power flow equations~\cite{kundur2007power}. However, by using the soft constraints approach for the loss function~\eqref{eq:loss-fn}, it may be difficult to satisfy the dynamic and algebraic equations exactly. This can be seen as follows. If the weight coefficients $w_f$ and $w_g$ are selected too large, which severely penalizes the violation of the DAEs, the optimization problem may become ill-conditioned, and, hence, it may be difficult to converge to a minimum. On the other hand, if the values selected for~$w_f$ and $w_g$ are too small, the solution will not satisfy the dynamic equations or will not lie in the manifold described by the algebraic equations.  

To impose the DAEs as approximate hard constraints, we implement the \textit{penalty-based method} introduced in~\cite{lu2021physics} and summarized in Algorithm~\ref{alg:penalty-method}. The main idea behind this method is to replace the optimization problem with equality constraints (\ie the differential and algebraic equations) with a \textit{sequence} of unconstrained problems with varying penalty coefficients~$w^k_f$ and $w^k_f$. More specifically, during the $k$th ``outer'' iteration, we solve the following unconstrained optimization problem
$$
\min_{\theta}~\mathcal{L}(\theta;\mathcal{T}) = w^k_f \mathcal{L}_f + w_f^k \mathcal{L}_g.
$$
where $w^k_f$ and $w^k_g$ are the penalty coefficients for the $k$th iteration. Furthermore, at the beginning of each iteration, we increase the penalty coefficients by a constant factor $\beta > 1$:
\begin{align*}
w^{k+1}_g &= \beta w^{k}_g = (\beta)^k w^{0}_g, \\
w^{k+1}_f &= \beta w^{k}_f = (\beta)^k w^{0}_f.
\end{align*}
As $k \to \infty$, and given that the neural networks are well trained, the solution of the sequence of unconstrained optimization problems will converge to the solution that satisfies the DAEs approximately as hard constraints~\cite{luenberger1973introduction,lu2021physics}. In practice, however, if we fail to carefully select the hyper-parameters $w^0_f$, $w^0_g$, and $\beta$, the optimization problem may become ill-conditioned or experience slow convergence. 
\begin{algorithm}[t]
\DontPrintSemicolon
\SetAlgoLined
\textbf{Hyperparameters:} initial penalty coefficients~$w^0_f$ and $w^0_g$, factor~$\beta$, and number of iterations~$K$\;
$k \longleftarrow 0$\;
$\theta^0 \longleftarrow$ $\text{argmin}_\theta~\mathcal{L}^0(\theta;\mathcal{T})$: train the neural network~\eqref{eq:loss-fn} from random initialization, until the training loss has converged, \ie $\mathcal{L}^0(\theta;\mathcal{T}) \le 1$e-5\;
 \While{$k \le K$}{
  $k \longleftarrow k+1$\;
  $w^k_g \longleftarrow \beta w_g^{k-1}$\;
  $w^k_f \longleftarrow \beta w_f^{k-1}$\;
  $\theta^k \longleftarrow$ $\text{argmin}_\theta~\mathcal{L}^k(\theta;\mathcal{T})$: train the networks~\eqref{eq:loss-fn} from the initialization $\theta^{k-1}$, until the training loss has converged, \ie $\mathcal{L}^k(\theta;\mathcal{T}) \le 1$e-5\.;
 }
 \caption{Training using the penalty method~\cite{lu2021physics}}
 \label{alg:penalty-method}
\end{algorithm}

\vspace{-1em}
\subsection{Simulating DAEs for long-time horizons}
\label{ssec:simulating-DAEs}
\begin{figure}[t!]
\centering
\includegraphics[width=.5\textwidth, height=5.0cm]{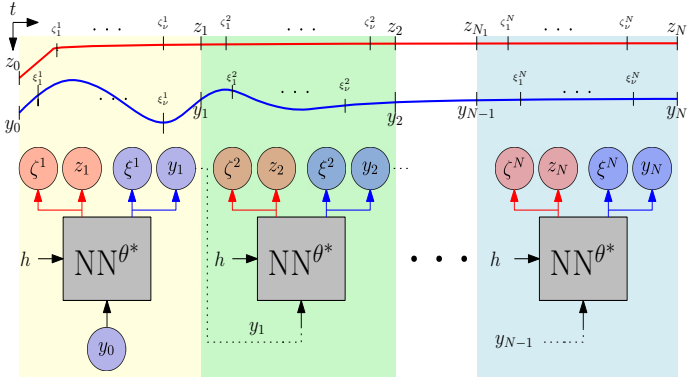}
\caption{Illustration of how to use the proposed trained DAE-PINN ($\text{NN}^{\theta^*}$) to simulate index-1 DAEs~\eqref{eq:DAE} for long-time horizons (\ie for $N$ time steps of size~$h$).}
\label{fig:NN-architecture}
\end{figure}

Until now, we have described how the DAE-PINN framework enables integrating DAEs~\eqref{eq:DAE} from $(t_n, y_n, z_n)$ to $(t_n+h,y_{n+1},z_{n+1})$.
Such a framework can use a large number $\nu$ of stages to take a very large time step~$h$. However, when simulating stiff and nonlinear DAEs (\eg the power network dynamics) for long-time horizons, it may be necessary to take multiple time steps. Thus, in this subsection, we briefly describe how we can use our trained DAE-PINN framework to simulate DAEs for long-time horizons. We provide a detailed description of the proposed iterative strategy in Algorithm~\ref{alg:integrating-long-time-horizons} and an illustrative example in Fig.~\ref{fig:NN-architecture}. The main idea behind our strategy is to update recurrently (in Markov-like fashion) the input to DAE-PINN~$y_n$ using the predicted dynamic states~$y^{\theta^*}_{n+1}$ from the previous evaluation step.

Thus, Algorithm~\ref{alg:integrating-long-time-horizons} enables us to simulate the solution trajectories of DAEs~\eqref{eq:DAE}, $y(t)$ and $z(t)$, within the time interval $t \in [0, h \cdot N]$, using a single trained DAE-PINN with time step~$h$. We remark that one can easily extend Algorithm~\ref{alg:integrating-long-time-horizons} to work with multiple trained discrete PINNs with possibly different time steps~$h$. Such a strategy can be applied, for example, to problems with multiple time scales (\eg transients and steady-state).

\begin{algorithm}[t]
\DontPrintSemicolon
\SetAlgoLined
Given is the number of time steps~$N$, and the DAE-PINN with \textit{trained} parameters $\theta^*$ and time step~$h$. Let the initial condition of~\eqref{eq:dynamic}, $y_0$, be the input to the DAE-PINN, \ie $y_n = y_0$. \;
 \For{k = 1,\ldots,N}{
  (1)~compute the forward pass using the proposed framework, \ie\;
  \vspace{-1em}
  \begin{align*}
      y_n &\mapsto [\xi^{\theta^*}_1, \ldots, \xi^{\theta^*}_\nu, y^{\theta^*}_{n+1}] =: Y^{\theta^*}_k \\
      y_n &\mapsto [\zeta^{\theta^*}_1,\ldots,\zeta^{\theta^*}_\nu,z^{\theta^*}_{n+1}] =: Z^{\theta^*}_k
  \end{align*}
  (2)~update the input using the predicted value $y_{n+1}^{\theta^*}$, \ie\;
  \vspace{-1em}
  \begin{align*}
      y_n \longleftarrow y_{n+1}^{\theta^*}
  \end{align*}
 }
 The Algorithm computes the solution trajectory in the time interval $[0, h \cdot N]$. Such a solution trajectory is obtained by concatenating the outputs from all forward passes, \ie $\{Y^{\theta^*}_k\}_{k=1}^N$ and $\{Z^{\theta^*}_k\}_{k=1}^N$ \;
 \caption{Simulating DAEs for long-time horizons}
 \label{alg:integrating-long-time-horizons}
\end{algorithm}

%% Proof of concept
\section{Numerical Experiments}
\label{sec:numerical-experiments}
\begin{figure}[t!]
\centering
\includegraphics[width=8cm, height=5.25cm]{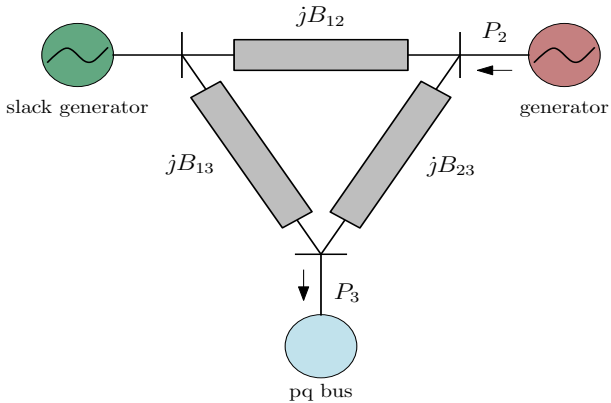}
\caption{Three-bus-two-generators power network~\cite{zheng2010bi}}
\label{fig:three-bus}
\vspace{-1em}
\end{figure}

This section contains a systematic study on a three-bus power (Fig.~\ref{fig:three-bus}) network that aims to demonstrate the performance of our DAE-PINN framework. 
\vspace{-1em}
\subsection{Three-bus power network} \label{ssec:3-bus-net-description}
We consider the three-bus (a slack bus, a generator bus, and a load bus) power network depicted in Fig.~\ref{fig:three-bus} and described by the following set of nonlinear and stiff DAEs~\cite{zheng2010bi}
\begin{subequations} \label{eq:power-network}
\begin{align} 
    \dot{\omega}_1 &= (1/M_1) (-D \omega_1 + f_1 + f_2) \\
    \dot{\omega}_2 &= (1/M_2) (-D \omega_2 - f_1) \\
    \dot{\delta}_2 &= \omega_2 - \omega_1 \\
    \dot{\delta}_3 &= - (\omega_1 - f_2/D_l) \\
    0 &= - (1/V_3)(g_1), \label{eq:power-network-algebraic}
\end{align}
\end{subequations}
where $y = (\omega_1, \omega_2, \delta_2, \delta_3)^\top$ are the dynamic states, $z = V_3$ the algebraic state, and 
\begin{align*}
    f_1 &= B_{12} V_1 V_2 \sin(\delta_2) +  B_{23} V_2 V_3 \sin(\delta_2 - \delta_3) + P_g,\\
    f_2 &= B_{13} V_1 V_3 \sin(\delta_3) +  B_{23} V_2 V_3 \sin(\delta_3 - \delta_2) + P_l,\\
    g_1 &= (B_{13} + B_{23}) V_3^2 - B_{13} V_1 V_3 \cos(\delta_3) \ldots \\
    &\qquad~\qquad~\qquad~\quad-  B_{23} V_2 V_3 \cos(\delta_3 - \delta_2) + Q_l.
\end{align*}
We fix the parameters of the power network to the following values $M_1 = .52$, $M_2 = .0531$, $D = .05$, $D_l = .005$, $V_1=1.02$, $V_2 = 0.05$, $B_{12}, B_{13}, B_{23} = 10$, $P_g = -2.0$, $P_l=3.0$, and $Q_l = .1$.
% ----------------------------
% Hyper-parameters and training protocols
\subsection{Neural Networks, hyper-parameters and learning protocols} \label{ssec:Hyperparamters}
We implemented DAE-PINN using PyTorch and published all the codes in GitHub. All the experiments presented in this section were trained by minimizing the loss function~$\mathcal{L(\theta;\mathcal{T})}$~\eqref{eq:loss-fn} using the Adam~\cite{kingma2014adam} optimizer with default hyper-parameters and initial learning rate $\eta = 10^{-3}$. We reduced the learning rate whenever the value of the loss function~$\mathcal{L}$ reached a plateau or started to increase. The training and test datasets consist of initial conditions collected uniformly at random and as follows: $\omega_1(0), \omega_2(0) \sim \mathcal{U}(-\pi,\pi)$ and $\delta_2(0), \delta_3(0) \sim \mathcal{U}(-0.1,0.1)$.

The neural network that approximates the mapping $y_n \mapsto (\xi_1,\ldots,\xi_\nu,y_{n+1})$ (\ie dynamic equations) and the neural network that approximates the mapping $y_n \mapsto (\zeta_1,\ldots,\zeta_\nu,z_{n+1})$ (\ie algebraic equations) were implemented using the improved fully-connected architecture proposed in~\cite{wang2020understanding}, which has the following forward pass:
\begin{align*}
&U = \phi(X W^1 + b^1),~V = \phi(X W^2 + b^2) \\
&H^{(1)} = \phi(X W^{z,1} + b^{z,1}) \\
&Z^{(k)} = \phi(H^{(k)} W^{z,k} + b^{z,k}),~k=1, \ldots,d \\
&H^{(k+1)} = (1 - Z^{(k)}) \odot U + Z^{(k)} \odot V,~k=1,\ldots,d \\
&f_\theta (x) = H^{(d+1)} W +b,
\end{align*}
Here, $X$ is the input tensor to the neural network, $d$ is the number of hidden-layers (\ie the network's depth), $\odot$ is the Hadamard or element-wise product, and $\phi$, in this paper, is a point-wise sinusoidal activation function. We also assume that each hidden-layer has width $w$. The trainable parameters of this novel network architecture, which we initialize using the Glorot normal algorithm, are collected in the following set:
$$
\theta = \{W^1, b^1, W^2, b^2, \{W^{z,l}, b^{z,l}\}_{l=1}^d, W, b\}.
$$
Our experiments show that this novel architecture outperforms the conventional fully connected architecture. This is because it explicitly accounts for the multiplicative interactions between different inputs and enhances hidden-state representation with residual connections~\cite{wang2020understanding}. Let us conclude this subsection with the following remark.
\begin{remark} 
\textit{Output feature layer for the algebraic equation.} The term $(1/V_3)$ in~\eqref{eq:power-network-algebraic} may lead to the loss for the algebraic variables~$\mathcal{L}_g$ being a few orders of magnitude larger than the loss for the dynamic variables~$\mathcal{L}_f$. We have observed empirically that such an imbalance compromises gradient descent optimization. To mitigate such an issue, we added the following output feature layer for the neural network associated with the algebraic equations:
$$
(\zeta_1,\ldots,\zeta_\nu,z_{n+1}) \mapsto \text{softplus}(\zeta_1,\ldots,\zeta_\nu,z_{n+1}).
$$
Note that the above feature layer constraints the bus voltage to be non-negative, \ie~$V_3>0$.
\end{remark}

% ---------------------------
% convergence tests
\subsection{Convergence experiments} \label{ssec:convergence-experiments}
In this subsection, we investigate how the network architecture, the network size, and the training dataset affect the training convergence of DAE-PINN. To this end, we train DAE-PINN with a time step $h=0.1$ for 50000 epochs. 

\subsubsection*{Network architecture} Our first convergence experiment evaluates which architecture, stacked or unstacked, provides better convergence results for the training of DAE-PINN. The stacked architecture uses a neural network (width $w = 25$ and depth $d=4$ layers) for each dynamic and algebraic state.  On the other hand, the unstacked architecture uses one neural network (width $w = 100$ and depth $d=4$ layers) for all the dynamic states and another neural network (width $w=25$ and depth $d=4$ layers) for the algebraic state. Fig.~\ref{fig:stacked-vs-unstacked} shows the results of running this experiment 10 times. We observe that the unstacked architecture provides us with the best training performance. 

\subsubsection*{Network size} This experiment evaluates the effect of the size of the networks during training. More specifically, we use an unstacked architecture (to eliminate the network architecture effect) to verify the training convergence while varying the width and depth of the neural networks. Fig.~\ref{fig:width} illustrates the results when we vary the width (depth fixed to $d=2$ layers). We note that increasing the width from 10 to 200 decreases the train and test errors, but the errors increase instead when we further increase the width. On the other hand, Fig.~\ref{fig:depth} shows the results when we vary the depth (width fixed to $w=100$). We observe that the train and test errors reach a minimum when we set the depth to $d=4$ or $d=8$ hidden layers. 

\subsubsection*{Training dataset} 
\begin{figure}
     \centering
     \begin{subfigure}[b]{0.45\textwidth}
         \centering
         \includegraphics[width=1.0\textwidth, height=5.25cm]{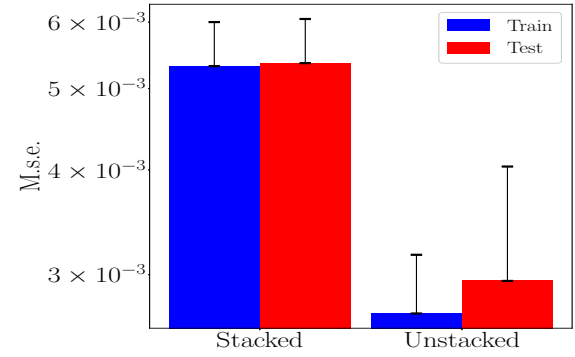}
         \caption{}
         \label{fig:stacked-vs-unstacked}
     \end{subfigure}
     \hfill
     \begin{subfigure}[b]{0.45\textwidth}
         \centering
         \includegraphics[width=1.0\textwidth, height=5.25cm]{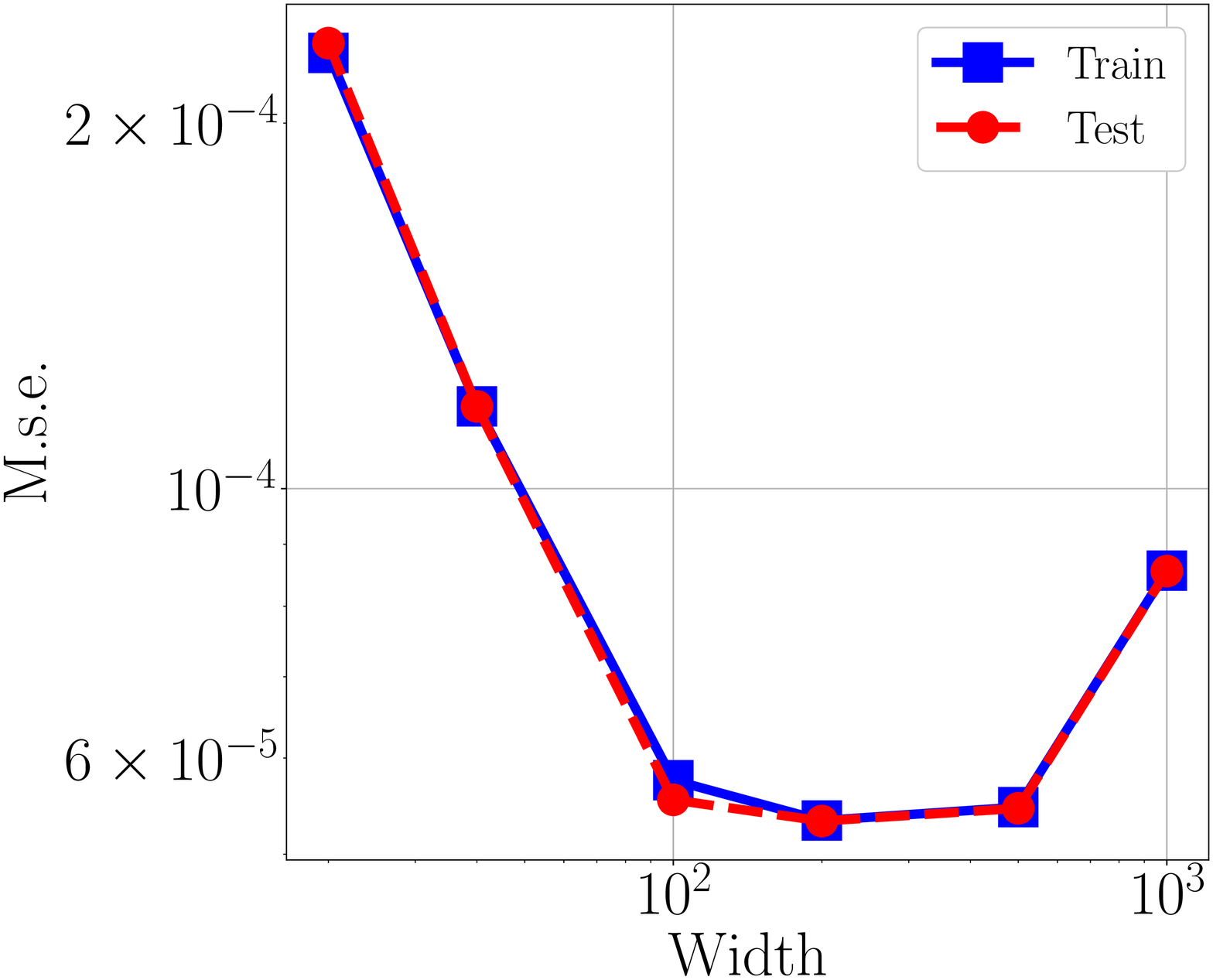}
         \caption{}
         \label{fig:width}
     \end{subfigure}
     \hfill
     \begin{subfigure}[b]{0.45\textwidth}
         \centering
         \includegraphics[width=1.0\textwidth, height=5.25cm]{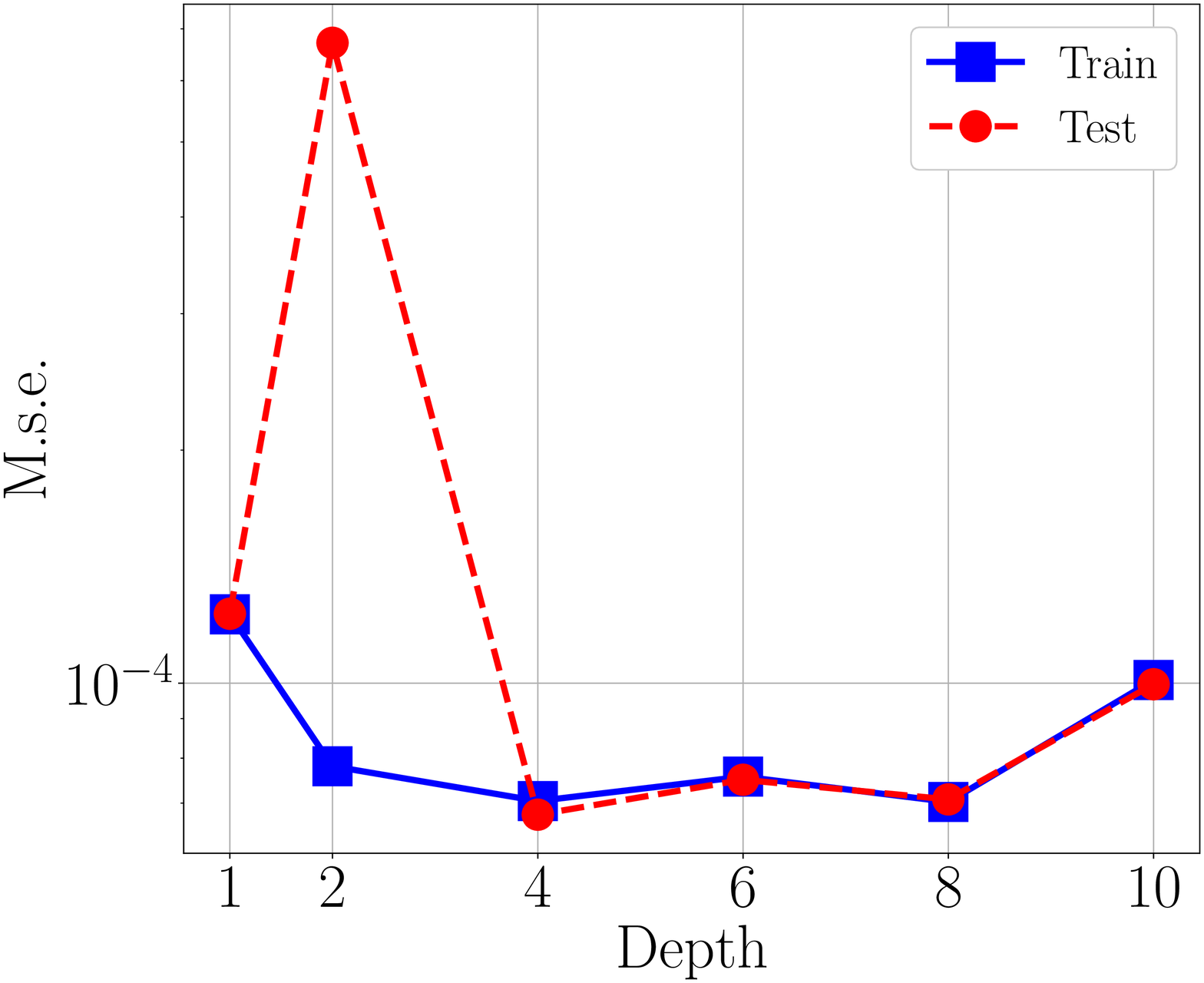}
         \caption{}
         \label{fig:depth}
     \end{subfigure}
     \hfill
     \begin{subfigure}[b]{0.47\textwidth}
         \centering
         \includegraphics[width=1.0\textwidth, height=5.25cm]{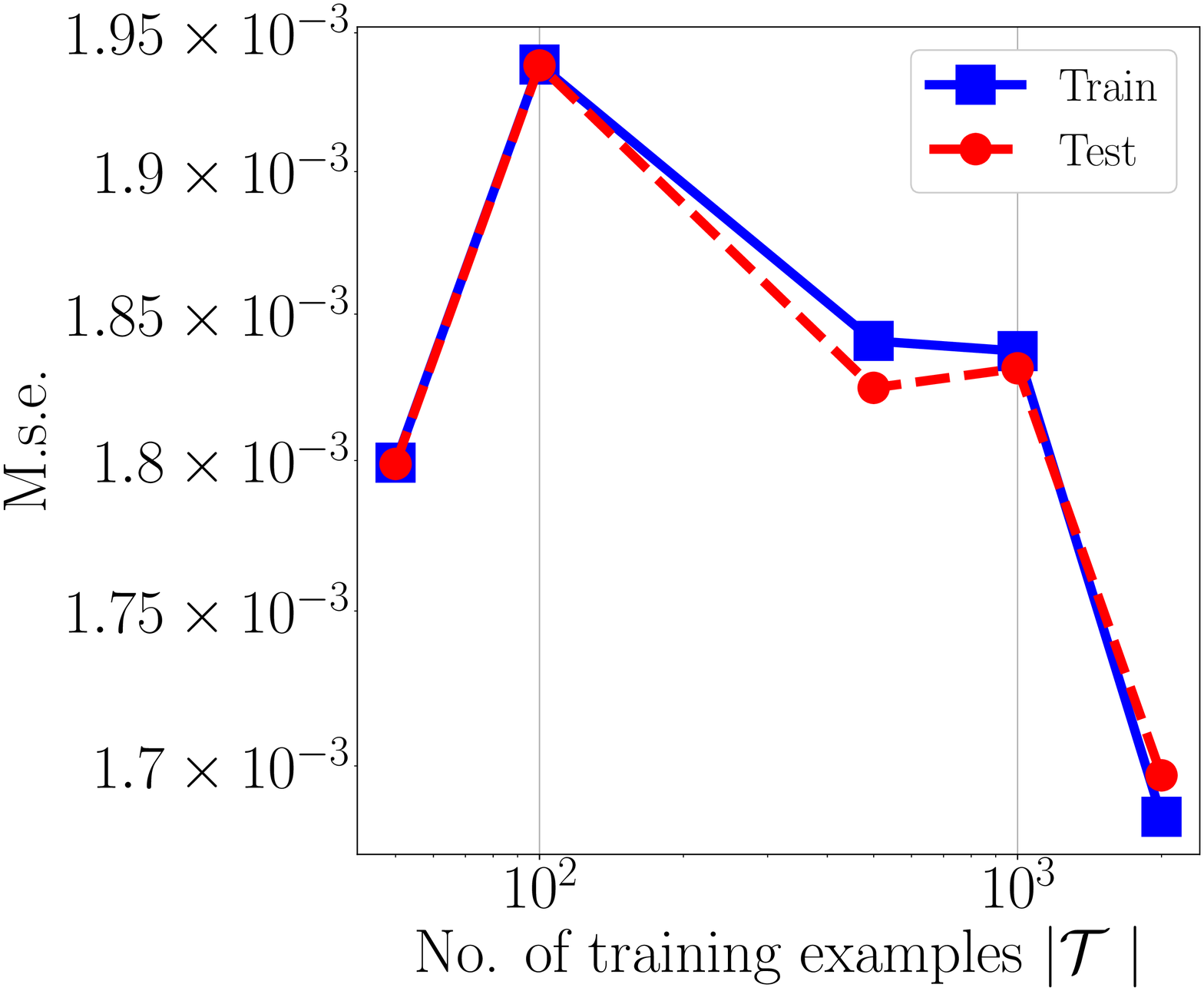}
         \caption{}
         \label{fig:num-train}
     \end{subfigure}
     \caption{Convergence experiments. (a) stacked vs unstacked architectures. (b) Network width. (c) Network depth. (d) Number of training examples.}
        \label{fig:convergence-test}
\end{figure}

Our last experiment investigates how the size of the training dataset~$|\mathcal{T}|$ affects the training convergence of DAE-PINN. To eliminate the effect of the architecture and size of the neural networks, we choose unstacked architectures with depth of $d = 4$ for the neural networks of the dynamic and algebraic states. Further, we select a width of $w=100$ (resp. $w=40$) for the neural network of the dynamic (resp. algebraic) states. The results illustrate that (see Fig.~\ref{fig:num-train}) including more training examples (initial conditions), in general, leads to smaller train and test errors. 
We conclude this section by describing the characteristics of our best DAE-PINN model. This best model trains with $|\mathcal{T}| = 2000$ initial condition points sampled from the state-space, tests the performance of DAE-PINN every $1000$ epochs using a test dataset witn $1500$ initial conditions not included in the train dataset, and uses unstacked neural network architectures with $d=4$ hidden layers. Moreover, for this best DAE-PINN model, the neural network representing the dynamic (resp. algebraic) states has a width of $w = 100$ (resp. $w=40$). 

% ---------------------------
% results for the best-trained model
\subsection{Results for the best DAE-PINN model} \label{ssec:best-trained-model}
In this subsection, we verify the effectiveness of DAE-PINN to perform long-time simulation of DAEs using Algorithm~\ref{alg:integrating-long-time-horizons}. To this end, we train the best DAE-PINN model with time-step $h=0.1$ using the penalty-method described in Algorithm~\ref{alg:penalty-method} with hyper-parameters $w_{f}^0 = w_g^0 = 1$ and $\beta = 2$. 

\begin{figure}[t!]
\centering
\includegraphics[width=.45\textwidth, height=10cm]{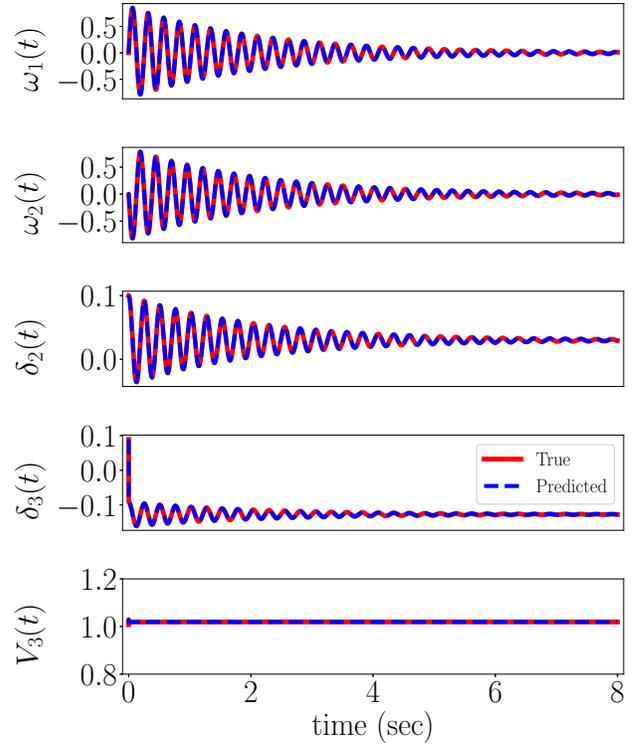}
\caption{Predicted and true solution trajectories of the DAEs describing the three-bus power network dynamics~\eqref{eq:power-network} within the simulation time interval $[0, N \cdot h] = [0,8]$ seconds for a initial condition sampled from the test dataset.}
\label{fig:predicted-true}
\end{figure}

Fig.~\ref{fig:predicted-true} presents a simulated DAE trajectory for $N=80$ time steps, corresponding to a representative initial condition selected uniformly at random from the test dataset. We note excellent agreement between the simulated trajectory and the true trajectory (obtained by integrating~\eqref{eq:power-network} using conventional numerical methods~\cite{wanner1996solving}). To better understand the long-time simulation accuracy of the DAE-PINN framework, we sample $100$ initial conditions from the test dataset and compute the mean and standard deviation of the $L^2$ relative error of each state variable. Table~\ref{table:mean-std} reports the $L^2$ relative errors of each state variable. From the reported results, we conclude that DAE-PINN can simulate DAEs for long-time horizons with excellent accuracy.

\begin{table}[h!]
\centering
\begin{tabular}{c|c c c c c} 
 \hline
  & $\omega_1$ & $\omega_2$ & $\delta_2$ & $\delta_3$ & $V_3$ \\  
 \hline\hline
 \textbf{mean} & 0.0382  & 0.0381  & 0.0093  & 0.0011 & 0.0002 \\ 
 \textbf{st. dev.} & 1.01$e$-2 & 1.07$e$-2 & 2.44 $e$-3 & 2.96$e$-4
 & 2.98$e$-07 \\
 \hline
\end{tabular}
\caption{Mean and standard deviation of the $L^2$ relative error of the long-time simulation of $100$ initial conditions sampled from the test dataset.}
\label{table:mean-std}
\end{table}
\vspace{-1em}
% ---------------------------
% results for the best-trained model
\subsection{Comparison with other numerical integration schemes} \label{ssec:comparison-other}
\begin{figure}[t!]
\centering
\includegraphics[width=.45\textwidth, height=10cm]{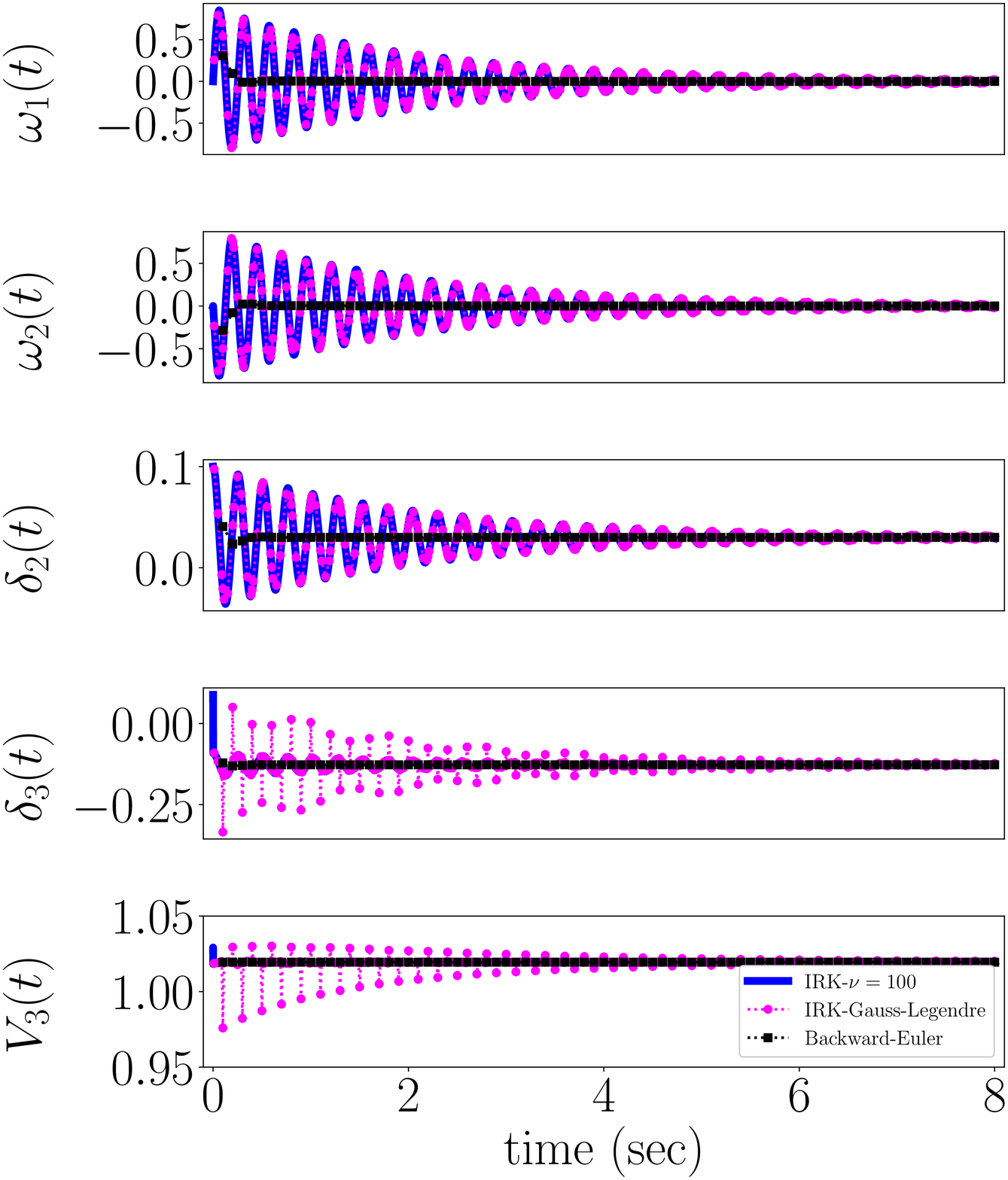}
\caption{Comparing the long-time simulation accuracy of DAE-PINNs enabling (i) IRK scheme with $\nu=100$ stages, (ii) IRK Gauss-Legendre scheme, and (iii) Backward-Euler method.}
\label{fig:comparison}
\vspace{-1em}
\end{figure}

\begin{figure}[t!]
\centering
\includegraphics[width=.45\textwidth, height=6cm]{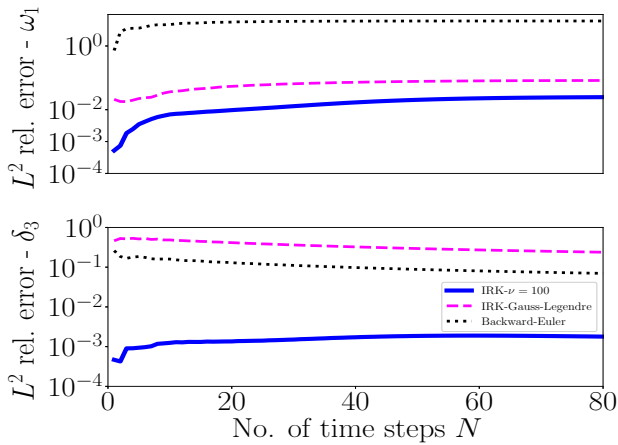}
\caption{$L^2$ relative error for the slack generator speed~$\omega_1$ and load bus angle~$\delta_3$ as a function of the number of time steps~$N$.}
\label{fig:comparison-l2-N}
\end{figure}

In this subsection, we compare the proposed DAE-PINN framework that enables the IRK scheme with $\nu = 100$ stages with other discrete PINN models enabling the following DAE numerical integration schemes~\cite{iserles2009first}: (i) Backward-Euler method and (ii) the Gauss-Legendre IRK with $\nu=3$, which is probably the largest IRK scheme that is consistent, stable, and with reasonable implementation costs~\cite{iserles2009first}. We train and implement all the previously mentioned discrete PINN models using the same protocols and with time-step $h=0.1$. We then test their capability of simulating DAEs for $N=80$ time steps. Fig.~\ref{fig:comparison} compares the three discrete PINN models for simulating the DAEs for the representative initial condition selected from the test dataset. The results clearly illustrate that our DAE-PINN, which enables the IRK scheme with $\nu=100$ stages, significantly outperforms all other discrete PINN models. We also illustrate (see Fig.~\ref{fig:comparison-l2-N}) the $L^2$ relative error as a function of the number of time steps~$N$ for the slack machine speed~$\omega_1$ and the load bus angle~$\delta_3$. One should observe that the discrete PINN model for the Gauss-Legendre IRK method effectively simulates the speed of the slack generator but fails to simulate the load bus angle. On the other hand, the discrete PINN model for the Backward-Euler method fails to simulate the machine speed and also the load bus angle dynamics.

% discussion
\section{Discussion} \label{sec:discussion}
\subsubsection*{On extending our framework to large-scale power networks}
Developing deep learning methods for simulating large-scale scientific and engineering systems remains an open problem. Thus, the straightforward application of DAE-PINN for simulating large-scale power networks is not feasible. We, however, believe that similar to the author's previous work~\cite{li2020machine}, our proposed framework can be used in a plug-and-play fashion and replace the numerical solvers for the DAEs describing the individual components of the network (e.g., generators). To showcase such plug-and-play ability, in our future work, we plan to construct a surrogate model that can predict the response of a medium-size power network and whose components are pre-trained using our framework. To this end, our method must generalize to unseen events and even predict unstable behaviors. The fully connected neural networks used in this paper may not be powerful enough for such a challenge. Thus, it is also part of our future work to enhance our method using more sophisticated architectures, which we briefly describe next. 

\subsubsection*{On using more sophisticated Neural Network architectures}
In this paper, to simulate DAEs over a long-time horizon, we employed a modified version of the conventional fully connected neural network architecture. However, we realize that other architectures may increase our ability to simulate long-time dependencies~\cite{zhou2020informer}. Thus, in our future work, we plan to implement DAE-PINN using neural networks that can generalize well to unseen events. In particular, we plan to employ the state-of-the-art deep Operator Neural Network~(deepOnet)~\cite{lu2021learning}, a neural network that approximates nonlinear operators (a mapping from functions to functions), which has shown great potential to reduce the generalization error significantly. We can apply deepOnets to our framework by noting that integration is an operation of the form: $T_h: x(\cdot) \mapsto x(\cdot + h)$ where the time-step $h$ is a parameter.

\subsubsection*{On the inverse problem}
We remark that extending the proposed framework to learn unknown but identifiable parameters of DAEs is straightforward (see~\cite{raissi2019physics} for more details). Furthermore, in \cite{misyris2020physics}, the authors already used a physics-informed continuous deep learning model to learn unknown parameters of the power network. It is, however, unclear whether the authors' framework learns the stiff nonlinear DAEs or a non-stiff ODE-based approximation of the power network dynamics. As reported in~\cite{ji2020stiff} (and also our experience with physics-informed continuous models), the learning process of stiff ODEs and DAEs using physics-informed continuous models is extremely unstable. Thus, it requires a problem-dependent solution to avoid the failure of gradient-based training. 

\subsubsection*{On the stochastic setting}
With the increasing penetration of renewable resources, the operating conditions for power networks are becoming more uncertain. Thus, developing an online dynamic security assessment tool that considers such a stochastic environment is necessary. To this end, in our future work, we will develop a deep learning framework that learns and simulates the stochastic differential-algebraic equations describing power networks dynamics for a given distribution of initial conditions and a set of uncertain parameters.

\section{Conclusion} \label{sec:conclusion}
We developed DAE-PINN, a deep learning framework for learning and simulating the set \textit{differential-algebraic equations}~(DAE) that describes power networks. DAE-PINN consists of a discrete \textit{physics-informed neural network} model that enables employing arbitrarily accurate implicit Runge-Kutta schemes with a large number of stages. Moreover, we implemented a penalty-based that enforces DAE-PINN to satisfy the DAEs as approximate hard constraints. We then proposed Algorithm~\ref{alg:integrating-long-time-horizons}, which uses the trained DAE-PINN to simulate DAEs over long-time horizons. Finally, we demonstrated the effectiveness of our proposed framework using a three-bus power network.

% \appendices

% \section{}

\section*{Acknowledgment}
The authors gratefully acknowledge the support of the National Science Foundation (DMS-1555072, DMS-1736364, DMS-2053746, and DMS-2134209), and Brookhaven National Laboratory Subcontract 382247, and U.S. Department of Energy (DOE) Office of Science Advanced Scientific Computing Research program DE-SC0021142).

\bibliographystyle{IEEEtran}
\bibliography{refs}

%\end{IEEEbiography}
%\newpage

\end{document}